\documentclass[conference]{IEEEtran}

\usepackage{amsmath} 
\usepackage{amssymb}  
\usepackage{subfig}
\usepackage{graphicx} 
\graphicspath{{images/}}
\usepackage{multicol}
\usepackage{mathtools}
\usepackage{color}
\usepackage{float}
\DeclarePairedDelimiter\floor{\lfloor}{\rfloor}

\makeatletter
\let\old@ps@headings\ps@headings
\let\old@ps@IEEEtitlepagestyle\ps@IEEEtitlepagestyle
\def\confheader#1{%
  \def\ps@headings{%
    \old@ps@headings%
    \def\@oddhead{\strut\hfill#1\hfill\strut}%
    \def\@evenhead{\strut\hfill#1\hfill\strut}%
  }%
  \def\ps@IEEEtitlepagestyle{%
    \old@ps@IEEEtitlepagestyle%
    \def\@oddhead{\strut\hfill#1\hfill\strut}%
    \def\@evenhead{\strut\hfill#1\hfill\strut}%
  }%
  \ps@headings%
}
\makeatother

\confheader{%
  2017 International Conference on Indoor Positioning and Indoor Navigation (IPIN), 18-21 September 2017, Sapporo, Japan
}

\IEEEoverridecommandlockouts
\IEEEpubid{\makebox[\columnwidth]%
       {\hfill 978-1-5090-6299-7/17/\$31.00 \textcopyright2017 IEEE}%
\hspace{\columnsep}\makebox[\columnwidth]{ }}

\renewcommand\IEEEkeywordsname{Keywords}

\begin{document}
%
\title{5-DoF Monocular Visual Localization \\ Over Grid Based Floor}


\author{\IEEEauthorblockN{Manash Pratim Das}
\IEEEauthorblockA{Department of Biotechnology \\
Indian Institute of Technology, \\ 
Kharagpur, WB, 721302, India \\
Email: mpdmanash@iitkgp.ac.in}
\and
\IEEEauthorblockN{Gaurav Gardi}
\IEEEauthorblockA{Department of Physics\\
Indian Institute of Technology, \\ 
Kharagpur, WB, 721302, India \\
Email: gaurav.gardi2@gmail.com}
\and
\IEEEauthorblockN{Jayanta Mukhopadhyay}
\IEEEauthorblockA{Department of Computer Science \\ and Engineering \\
Indian Institute of Technology, \\ 
Kharagpur, WB, 721302, India \\
Email: jay@cse.iitkgp.ernet.in}}

 


\maketitle

\begin{abstract}
Reliable localization is one of the most important parts of an MAV system. Localization in an indoor GPS-denied environment is a relatively difficult problem. Current vision based algorithms track optical features to calculate odometry. We present a novel localization method which can be applied in an environment having orthogonal sets of equally spaced lines to form a grid. With the help of a monocular camera and using the properties of the grid-lines below, the MAV is localized inside each sub-cell of the grid and consequently over the entire grid for a relative localization over the grid.

We demonstrate the effectiveness of our system onboard a customized MAV platform. The experimental results show that our method provides accurate 5-DoF localization over grid lines and it can be performed in real-time.
\end{abstract}


%
\IEEEpeerreviewmaketitle

\section{INTRODUCTION}

On-line indoor localization of Micro Aerial Vehicles (MAVs) in GPS-denied environment is performed by point cloud based scanning laser rangefinders and feature-based visual SLAM \cite{durrant2006simultaneous} in most cases. Extensive work has been done in the related fields leading to highly accurate localization methods. However, such localization techniques suffer in cases like large indoor open spaces where objects are sufficiently far enough to 1) be detected by light-weight MAV based laser range-finders, and 2) be considered as good features to be tracked for visual odometry. The recent mission 7 at International Aerial Robotics Competition (IARC) \cite{iarc} also challenges to demonstrate a new feature: ``navigation in a sterile environment with no external navigation aids such as GPS or large stationary points of reference such as walls." Localization methods as discussed above have already been tested in previous missions of IARC. The mission 7 of IARC features a floor with square grid cells. With the onset of indoor drones for industrial applications, we consider a system with low weight, cost, power consumption and on-board computation that consists of a single downward facing camera and a grid-based floor to address the challenge of indoor localization over large open spaces. 

We propose a monocular visual localization over grid-lines (mLOG) algorithm for indoor localization of MAVs that is accurate and computationally fast for real-time on-board processing. Our algorithm explicitly models the grid-lines and uses probabilistic clustering and labeling method to fit observed grid-lines to the model. A Random sample consensus (RANSAC \cite{fischler1981random}) method is used to detect outliers and reject the false positive lines before fitting the model. mLOG performs a five degree of freedom (5DoF) localization (position along X, Y, Z axis, roll and pitch) relative to the grid-based floor in a two-step sequential process. The first step involves localizing the MAV within a unit grid cell. Since a grid is a 2D plane of repeating unit cells (rectangles), the unit cells cannot be differentiated from each other when only a partial grid is visible. Hence, the relative positions are integrated using a winner take all (WTA) method to determine the position estimate over the grid-based floor in the second stage.

\subsection{Conventions and Assumptions}
\label{ConventionsAndAssumptions}
\begin{itemize}
	\item Image coordinates have their origin in the top-leftmost corner of the image, with positive $X_\text{I}$ axis to the right and positive $Y_\text{I}$ axis towards the  bottom.
	\item Coordinate system on the MAV is according to the Aircraft's Principle Axis convention.
	\item World coordinate system on the grid-based floor is according to right hand co-ordinate system with $X_\text{W}$ and $Y_\text{W}$ axis being forward and leftward respectively.
	\item The camera mounted on MAV is downward facing and aligned with the grid-lines on the floor, such that yaw($\gamma$) rotation taken with respect to the gridlines is zero. The roll($\alpha$) and pitch($\beta$) angles are taken with respect to the world frame. 
	\item The MAV speed is limited by the frame rate of the camera given by 
	\begin{equation*}
		v_{\text{max}} = \frac{m \times \epsilon_{\text{fps}}}{\epsilon_{\text{s}}} \textrm{m/s}
	\end{equation*}
	where $\epsilon_{\text{fps}}$ is the frames per second of the camera and $\epsilon_{\text{s}}$ is the speed factor. The speed factor $\epsilon_{\text{s}} \geq 3$ is the minimum number of frames that is desired while traversing over one unit cell length. We set $\epsilon_{\text{s}} = 3$ for best performance.
\end{itemize}

\section{RELATED WORK}
Works related to localization of MAVs can be classified into two parts: vision based and non-vision based in both indoor and outdoor environments. Compared to outdoor environments, localization is more difficult in indoor environments due to the unavailability of a global method like GPS. As discussed in the following, numerous work has been done on the localization of MAVs in indoor GPS-denied environments.

A vision based technique \cite{bimav} makes use of downward facing camera and visual odometry. RGBD camera and stereo vision are used in \cite{bachrach2011visual} for 3D localization of a MAV. In \cite{fang2015real}  RGBD camera is used in the estimation of depth to work in visually degraded environments. Use of beacons in \cite{leonard1991mobile} is also one of the methods tried for the purpose. The technique reported in \cite{artieda2009visual} makes use of visual 3-D SLAM for localization. Monocular visual odometry based method as proposed in \cite{forster2014svo} suffers from the limitation of the estimate being o	n an unknown scale. This problem is solved in \cite{weiss2012real} by estimating the scale with the data on an onboard inertial measurement unit (IMU).

 There are also a few non-vision based techniques reported in the literature. In \cite{bachrach2011range}, laser scan matching is used for pose estimation. Authors in \cite{hahnel2003learning} also use laser range finder for pose estimation and SLAM. In \cite{roberts2007quadrotor} infrared and ultrasonic sensors are used for indoor localization inside a room but the technique is not effective for long range sensing. The method reported in \cite{beul2015high} uses 3 stereo pairs, dual laser scanner and IMU for pose estimation. In \cite{mautz2012indoor} various indoor positioning systems are discussed. The technique reported in \cite{achtelik2009stereo} compares the two methods based on laser scanner and stereo vision. 
 The methods mentioned above fails to work efficiently in environments which have open spaces and the objects are far from the range of light weight MAV based laser scanners. Also, a method using downward facing camera fails when there is a uniform pattern on the floor.

 We propose a method which makes use of only monocular camera and no other sensors for inputs. In summary, the main contributions of this paper in relation to other work are:
\begin{itemize}
	\item An algorithm for consensus-based filtering of the set of detected lines from an image, such that the grid-lines are selected as inlier and other detected lines become outliers. The algorithm further clusters the inliers to group multiple lines detected from a single grid-line present in the image. (Section \ref{filteringAndClustering})
	\item An orientation estimation algorithm to estimate the roll and pitch angles of the camera from the detected grid-lines, and correct for the drift, caused by the angles in the detected grid-lines. (Section \ref{orientationAndCompensation})
	\item An algorithm to explicitly define a model of grid-lines and generate a cost function to fit the model on detected grid-lines for sub-cell localization. (Section \ref{subCellLocalization})
	\item A position integration algorithm to integrate the sub-cell position outputs from sequential image frames to determine the location of the MAV, relative to an initial position on the grid floor. (Section \ref{GridLocalization})
	\item Experimental results showing the accuracy and reliability of the proposed localization method in long run autonomous trials of the MAV. (Section \ref{exprerimentalResults})
\end{itemize}

\section{LINE FILTERING AND CLUSTERING}
\label{filteringAndClustering}

In the following sections, we assume that the grid-based floor is visible in the image. The set of lines detected from such an image might have lines other than those which belong to the grid-lines. Hence we filter the detected lines and cluster multiple detections before feeding them to model fitting algorithm (section \ref{subCellLocalization}).

\subsection{Detection and Representation of Lines} mLOG is independent of the method to detect lines from an image. In our implementation, we used OpenCV's implementation of Hough Transform. Since we can filter out the false positives (outliers), we use Hough Transform with threshold parameters that allowed for more false positives than false negatives. Each line is represented by a two-element ordered set  $(\rho, \theta)$. $\rho$ is the perpendicular distance between the line and the coordinate origin  $(0,0)$ (top-left corner of the image) in pixels. While $\theta$ is the angle in radians, the normal to the line makes with the $X$ axis of image. (i.e. $0 \sim \textrm{vertical line}, \pi/2 \sim \textrm{horizontal line} $). Let \(L_{\text{raw}}=\{\,(\rho,\theta) \in \mathbb{R}^2 \mid \rho \geq 0, -\pi \leq \theta < \pi\,\} \) be the set of all the detected lines. $L_{\text{raw}} = L_{\text{inliers}} \cup L_{\text{outliers}}$, where $L_{\text{inlier}}$ is a set of lines that belong to grid-lines, and $L_{\text{outlier}}$ is the set of lines which are not a part of the grid-lines, as detected from the image. Hence a line is represented by a point in $\rho,\theta$ space.

\begin{figure}[h]
    \vspace{-1.5cm}
	\subfloat{\def\svgwidth{0.5\columnwidth}{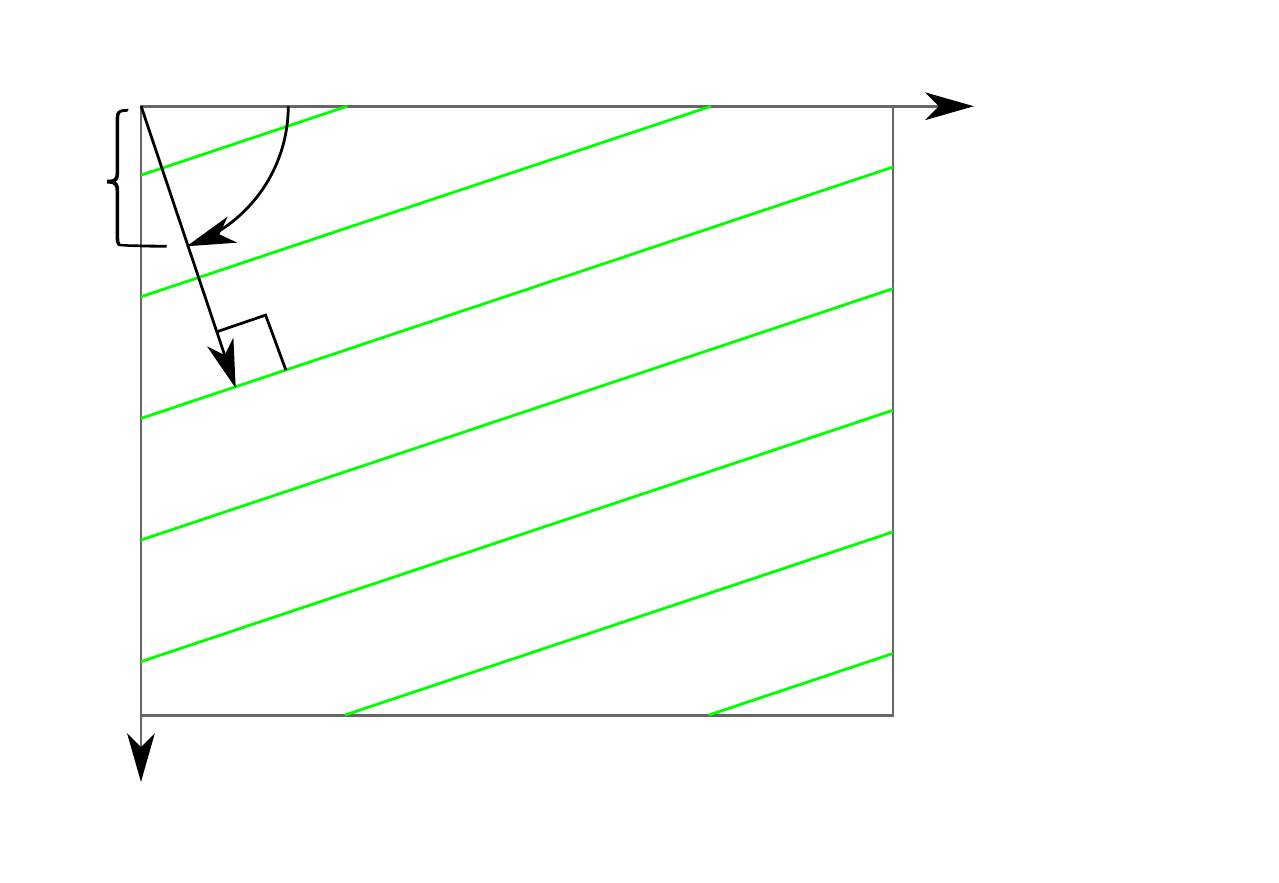}}
	\subfloat{\def\svgwidth{0.5\columnwidth}{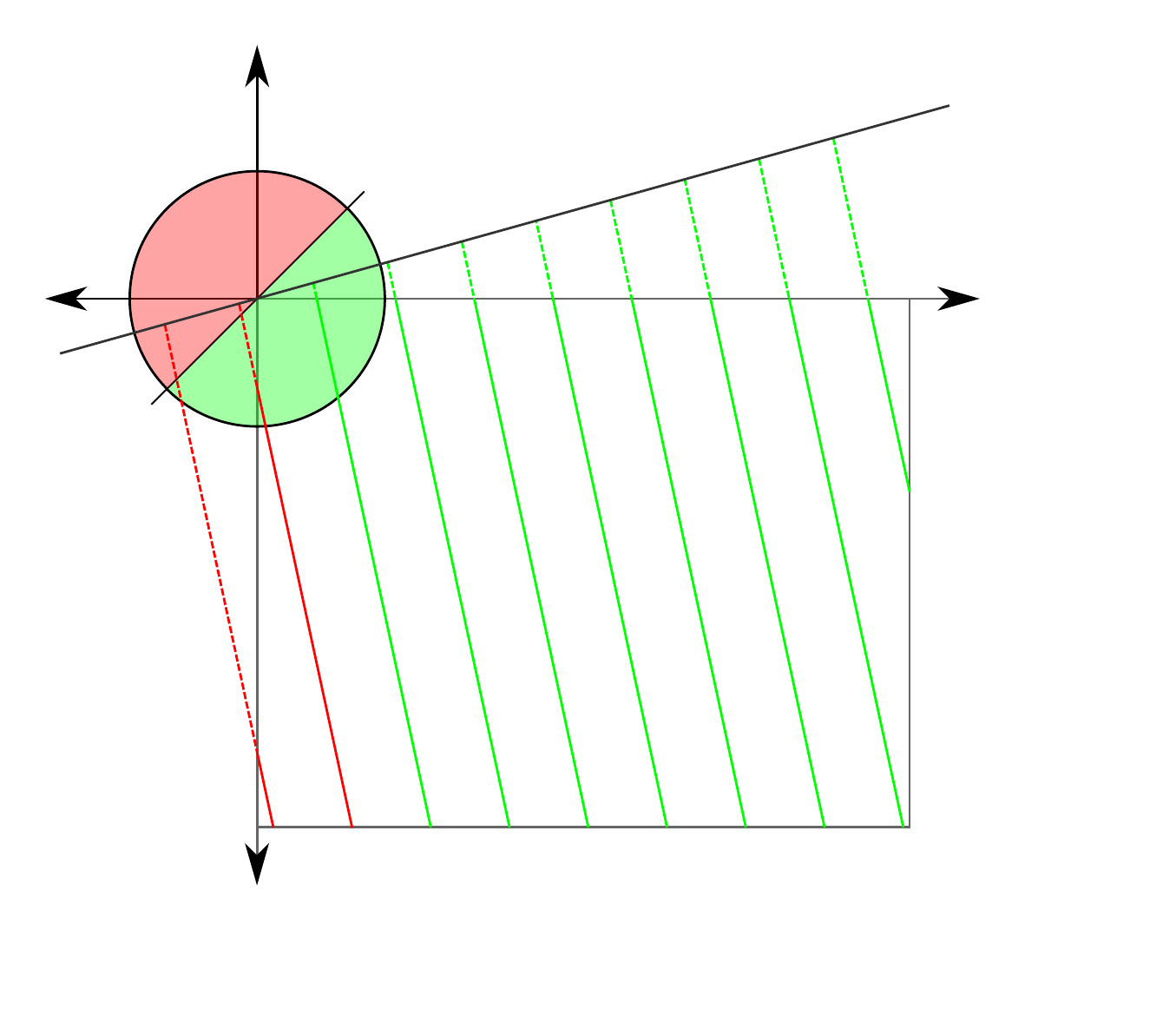}}
	\vspace{-0.7cm}
	\subfloat{\def\svgwidth{\columnwidth}{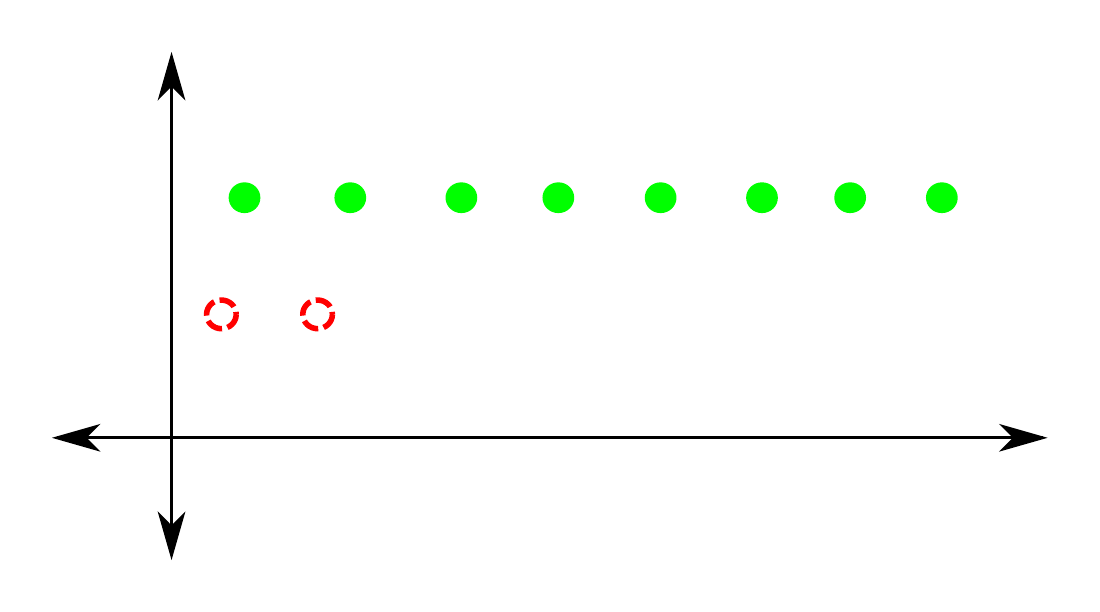}}
	\vspace{-0.2cm}
	\caption{\textbf{Top left}: Representation of a line using $\rho$ and $\theta$. \textbf{Top right}: Rejection of lines based on their $\theta$. The green region shows the allowed region, whereas the red region shows the rejected region. \textbf{Bottom}: Lines represented in $\rho,\theta$ space with discontinuity in the curve joining them. Here lines with color ``red (dotted circle)" are rejected since their respective $\theta$ lies in the rejected region.}
	\label{fig:rep&AngleRest}	
	\vspace{-0.2cm}
\end{figure}

\subsection{Filtering}
There are some orientations, in which the curve drawn in the $\rho,\theta$ space by joining the set of parallel lines of the grid (in order of appearance), becomes discontinuous. To remove such discontinuities, we reject all the ordered sets that does not meet the condition $-\pi/4 \geq \theta \geq \pi\times3/4$ (figure \ref{fig:rep&AngleRest})

\begin{figure}[h]
	\vspace{-0.1cm}
	\subfloat{\includegraphics[width=0.47\columnwidth]{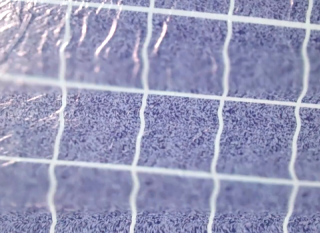}}
	\hspace{0.08cm}
	\subfloat{\includegraphics[width=0.47\columnwidth]{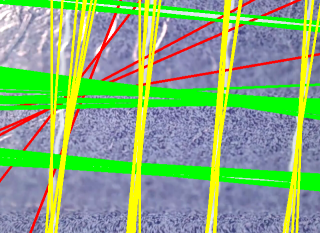}}\\
	\subfloat{\includegraphics[width=\columnwidth,trim={0 0 0 0.93cm},clip]{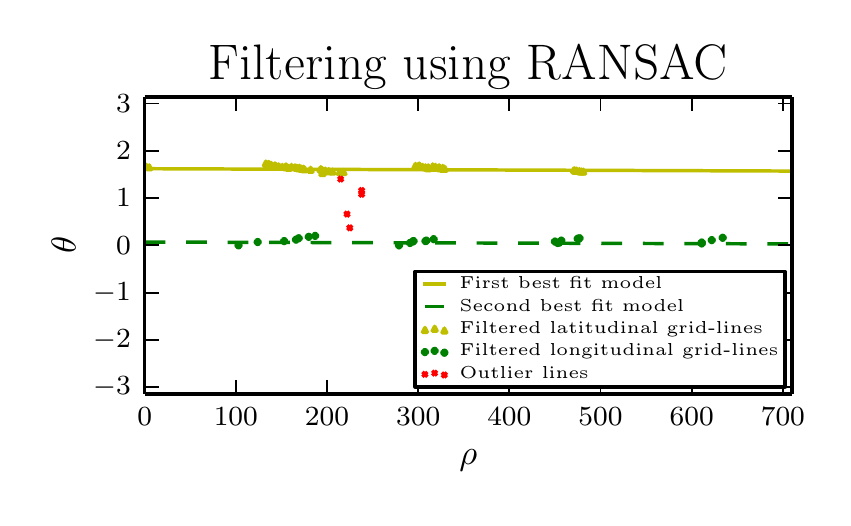}}
	\vspace{-0.4cm}
	\caption{Filtering of grid-lines using RANSAC. \textbf{Top left:} Input image from downward facing camera. \textbf{Top right:} Detected lines overlayed on the image. Red lines are recognised as outlier where as green and yellow lines as inlier after filtering. \textbf{Bottom:} $\theta$ vs $\rho$ graph where each line is represented by a point ($\theta,\rho$). Yellow and Green lines are the two best fit lines obtained by considering yellow and green points respectively as inlier and red points as outlier.}
	\label{fig:filtering}
	\vspace{-0.4cm}
\end{figure}

We further exploit the linear relationship between $\rho$ and $\theta$ for a set of parallel lines, and perform a Random Sampling Consensus (RANSAC) with a 2D linear model on the set of detected lines $L_{\text{raw}}$. RANSAC is performed twice without replacement to get two best fit inliers to the linear model, hence two best sets of parallel lines from $L_{\text{raw}}$. Further, the set of parallel lines (ordered set of $(\rho,\theta)$), with arithmetic mean of $\theta$ closer to $0$ is denoted as $L_{\text{long}}$ and that closer to $\pi/2$ is denoted as $L_{\text{lat}}$. Hence, the filtered set of lines $L_{\text{fil}}$, that contains only those detected lines which belong to the grid-lines (two sets of parallel lines with separation of around $\pi/2$ in mean $\theta$) is generated as $L_{\text{fil}} = L_{\text{long}} \cup L_{\text{lat}}$ (figure \ref{fig:filtering}). For RANSAC we empirically use threshold $d = 0.1$ for determining when a data point fits a model and $t = 10$ as the minimum number of close data values required to assert that a model fits well to data.

\subsection{Clustering}
\begin{figure}[h]
	\centering	
	\subfloat{\includegraphics[trim={0 0 0 0.93cm},clip]{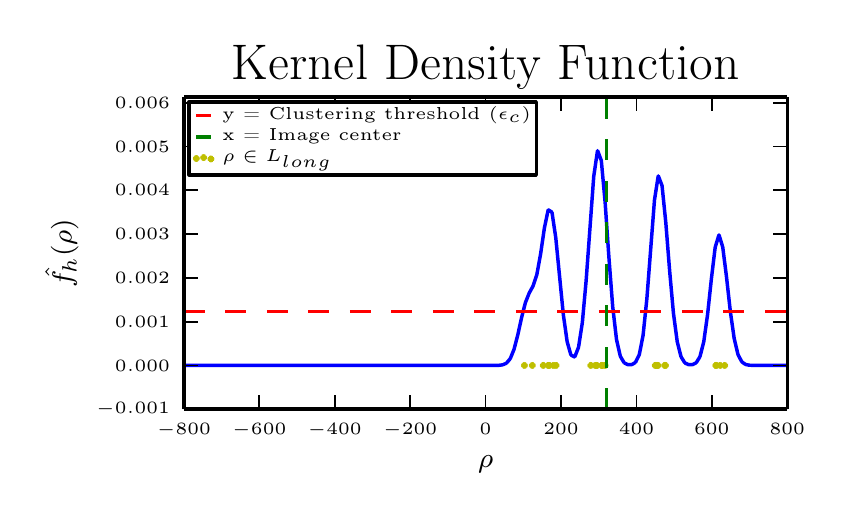}}
	\vspace{-0.2cm}
	\subfloat{\includegraphics[trim={0 0 0 0.93cm},clip]{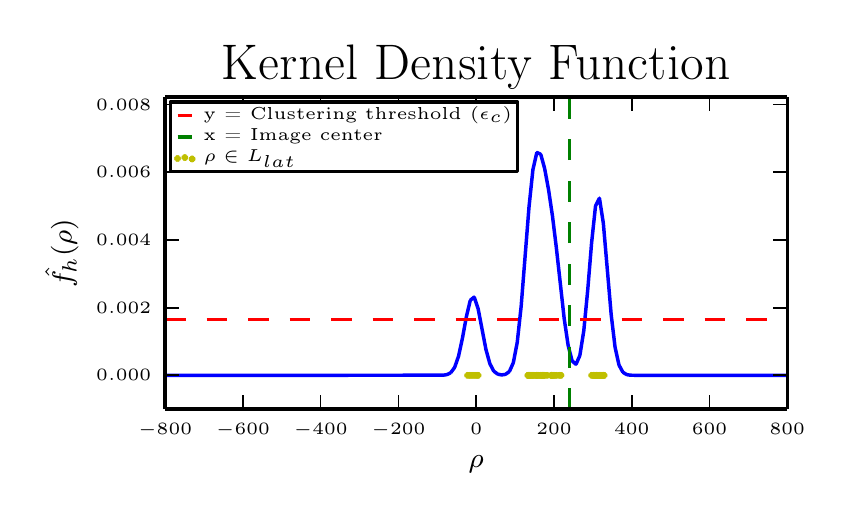}}
	\vspace{-0.4cm}
	\caption{Kernel Density Function and Clustering performed on set of longitudinal lines (\textbf{Top}) and latitudinal lines (\textbf{Bottom}).}
	\label{fig:kde}
	
\end{figure}

Let $L_{\text{real}}$ be the set of real world grid-lines that are visible in a particular image. There exists a non-injective and non-surjective relation between $L_{\text{real}}$ and $L_{\text{fil}}$. By clustering, we group the ``many to one'' lines to ``one to one'' (injective) lines with the mean $\rho$ and mean $\theta$ of each cluster and produce the set $L_{\text{inj}}$. We process the sets $L_{\text{lat}}$ and $L_{\text{long}}$ individually and consider only the $\rho$ values of the ordered set $(\rho,\theta)$ for clustering. We use univariate kernel density estimation (KDE) to estimate the probability density of $\rho$ in $L_{\text{lat}}$ and $L_{\text{long}}$ individually, as given by
\begin{equation}
\hat{f_b}(\rho) = \frac{1}{nb}\sum_{i=1}^{n}K\left( \frac{\rho-\rho_i}{b} \right)
\end{equation}
where $K$ is the standard normal probability density function
\begin{equation}
K(x) = \frac{e^{-\frac{1}{2}x^2}}{\sqrt{2\pi}}
\end{equation}
$n$ is the number of samples and $b$ is called the bandwidth which we empirically set as $b=20$ according to our data.
We hereby clip the space on $\rho$ axis at the regions of local minimums below a clustering threshold $\epsilon_c$ and form clusters. We empirically set $\epsilon_c = max \hat{f_b}(\rho) \times 0.25$ (figure \ref{fig:kde}).  

\begin{figure}[h]
	
	\subfloat{\includegraphics[width=0.47\columnwidth]{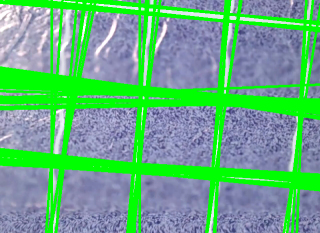}}
	\hspace{0.1cm}
	\subfloat{\includegraphics[width=0.47\columnwidth]{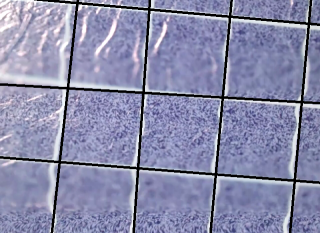}} 
	\caption{Clustering of grid-lines using KDE. \textbf{Left:} Input set of filtered lines ($L_{\text{fil}}$). \textbf{Right:} Set of clustered lines ($L_{\text{inj}}$) with mean $\rho$ and $\theta$ as output (black lines).}
	\label{fig:clustering}
\end{figure}

Let $\bar\rho^j$ and $\bar\theta^j$ be the means of the elements in $j^{th}$ cluster. The ordered set $(\bar\rho^j, \bar\theta^j)$ is sorted in descending order of $\bar\rho$ to form sets $L'_{\text{lat}}$ and $L'_{\text{long}}$, corresponding to latitudinal and longitudinal lines respectively. $L_{\text{inj}} = L'_{\text{lat}} \cup L'_{\text{long}}$ is the set of lines formed from $L_{\text{fil}}$ after clustering (figure \ref{fig:clustering})

\section{ORIENTATION ESTIMATION AND DRIFT CORRECTION}
\label{orientationAndCompensation}
The MAV's motion over the grid-based floor involves roll($\alpha$) and pitch($\beta$) rotations which change the view angle for the downward facing camera. As described in section \ref{subCellLocalization}, mLOG performs sub-cell localization based on the distance of the detected lines in the image from the image center, hence a change in view angle of the downward facing camera will lead to a drift in the detected lines. We therefore, estimate the $\alpha$ and $\beta$ angles of the camera to correct for the drift.

\subsection{Orientation Estimation}
\begin{figure}[h]
	\vspace{-0.7cm}
	\def\svgwidth{\columnwidth}{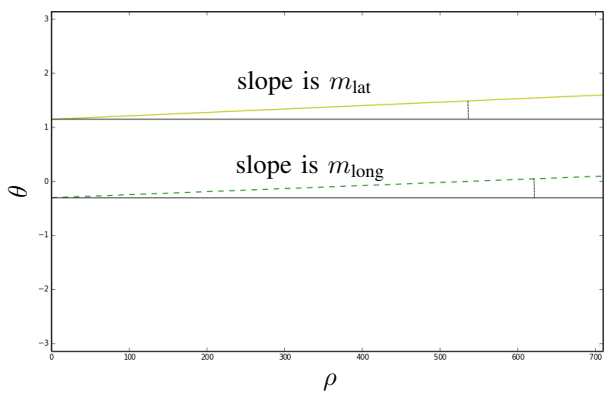}
	\vspace{-0.7cm}
	\caption{Slope of the line joining the ($\rho,\theta$) points of $L_{\text{lat}}$ and $L_{\text{long}}$, gives the $\alpha$ and $\beta$ angles.}
	\vspace{-0.2cm}
	\label{fig:orientEstima}
\end{figure}

Since we assume the yaw($\gamma$) rotation of the MAV to be small (section \ref{ConventionsAndAssumptions}), hence we only estimate the roll($\alpha$) and pitch($\beta$) rotation angles of the camera. In the $\rho,\theta$ space, the slopes of the linear curves $m_{\text{lat}}$ and $m_{\text{long}}$ (figure \ref{fig:orientEstima}), joining the ordered sets of parallel lines in $L_{\text{lat}}$ and $L_{\text{long}}$ are related to $\alpha$ and $\beta$ respectively as
\begin{equation}
	\alpha = tan^{-1}(m_{\text{lat}}) \times \epsilon_{\alpha}+\epsilon_{c\alpha}
\end{equation}
\begin{equation}
	\beta = tan^{-1}(m_{\text{long}}) \times \epsilon_{\beta}+\epsilon_{c\beta}
\end{equation}
where $\epsilon_{\alpha}$, $\epsilon_{\beta}$, $\epsilon_{c\alpha}$ and $\epsilon_{c\beta}$ are constants that can be determined by calibration with the ground truth values of $\alpha$ and $\beta$.

\subsection{Drift Correction and Labelling}
The $\alpha$ and $\beta$ rotations in the camera lead to drifts in the projection of grid-based floor in the image. If these angles are known, they can be used to correct for the drift such that the distance of detected grid-lines from the image center is independent of these rotation angles. Drift correction is performed by shifting the detected gridlines in $X_\text{I}$ and $Y_\text{I}$ axes by $x_{\text{drift}}$ and $y_{\text{drift}}$ pixels respectively, as given by
\begin{equation*}
\begin{aligned}
	x_{\text{drift}} = tan(\beta) \times f
\end{aligned}  \; \text{and} \;
\begin{aligned}
	y_{\text{drift}} = tan(\alpha) \times f
\end{aligned}
\end{equation*}
where $f$ is the focal length of the camera in pixels.

\begin{figure}[h]
	\vspace{-0.2cm}
	\subfloat{\includegraphics[width=0.5\columnwidth]{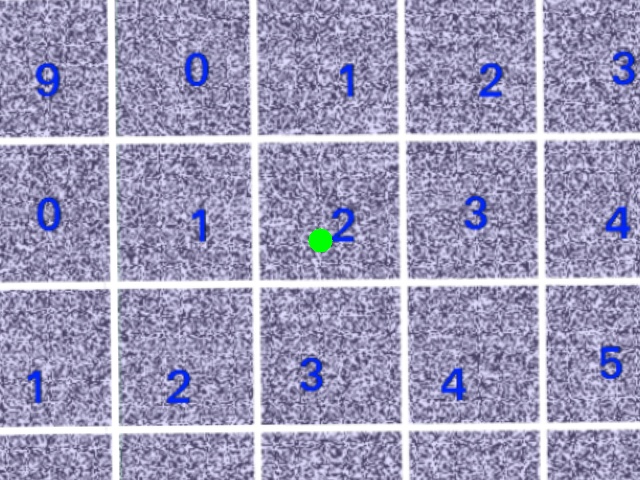}}
	\subfloat{\includegraphics[width=0.5\columnwidth]{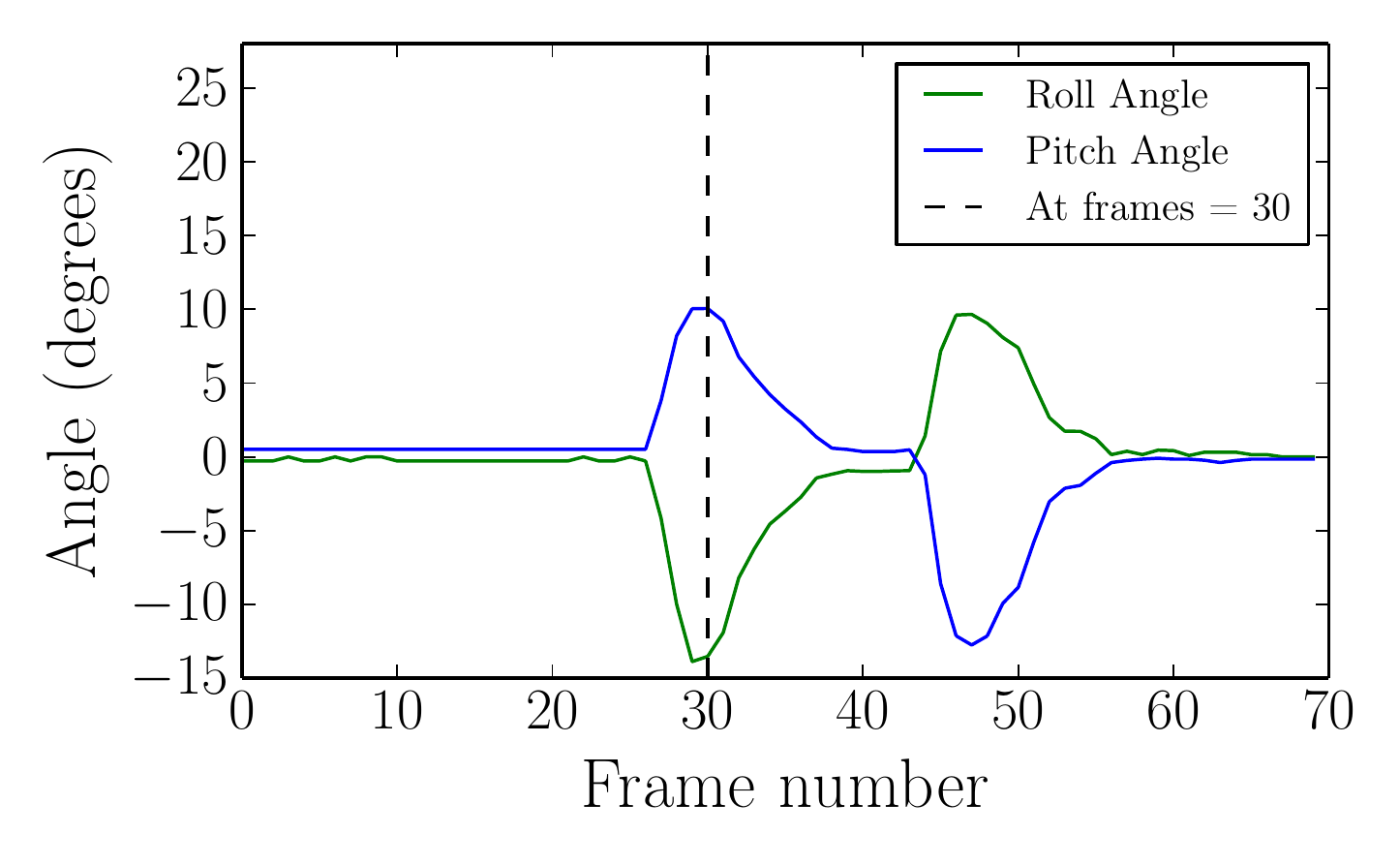}}\\
	\subfloat{\includegraphics[width=0.47\columnwidth]{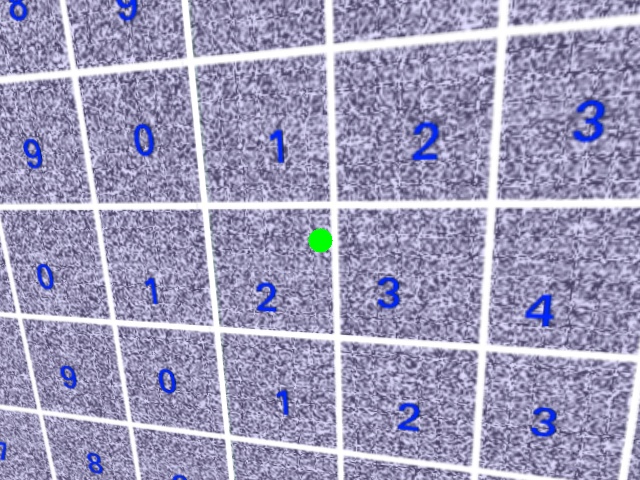}}
	\hspace{0.1cm}
	\subfloat{\includegraphics[width=0.47\columnwidth]{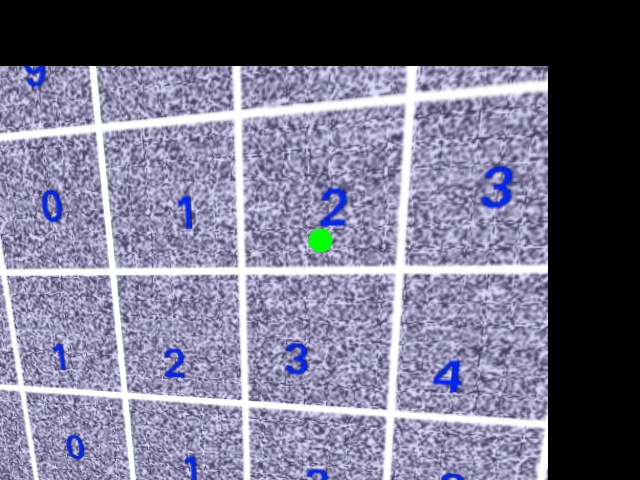}}
	\caption{\textbf{Top left}: Image from MAV while in steady state. \textbf{Top right}: Rotation angle profile of the MAV's movement. Consider the $30^\text{th}$ frame where MAV undergoes maximum roll and pitch rotation. \textbf{Bottom left}: Without drift correction, we can see that there has been significant amount of drift in grid's projection with respect to the image center (green dot). \textbf{Bottom right}: With drift correction, the projection has negligible drift.}
	\label{fig:driftCorrection}
	\vspace{-0.4cm}
\end{figure}

The ordered sets in $L'_{\text{lat}}$ and $L'_{\text{long}}$ are updated to $(\rho', \theta')$ for the lines after drift correction. As these two ordered sets were already sorted in descending order (in section \ref{filteringAndClustering}), they are further labelled according to their index of appearance shifted such that the ordered set to the top and left for $L'_{\text{lat}}$ and $L'_{\text{long}}$ respectively of the image center has label $l = 0$. The labelling for $L'_{\text{lat}}$ and $L'_{\text{long}}$ are independent of each other. According to the assumed convention for world (W) and image (I) co-ordinate frames, $Y_\text{W} = - X_\text{I}$ and $X_\text{W} = - Y_\text{I}$, the labelling is done such that it increases in the direction of increasing $X_\text{W}$ and $Y_\text{W}$. Thus the ordered sets of $L'_{\text{lat}}$ and $L'_{\text{long}}$ are updated to $(\rho', \theta', \l)$. Now, let $L_{\text{drift}} = L'_{\text{lat}} \cup L'_{\text{long}}$ be the set of lines formed from $L_{\text{inj}}$ after drift correction and labelling (figure \ref{fig:driftCorrection}).

\section{SUB CELL LOCALIZATION}
\label{subCellLocalization}
In a particular image with a partial view of the complete grid the position of the observed unit cell in the whole grid cannot be identified. Hence we initially determine the position only within the unit cell directly below the camera and integrate it later (section \ref{GridLocalization}). The model we define in mLOG considers the unit cell directly below the camera as the reference.

We use the $\rho'$ value in the ordered set $(\rho', \theta', \l)$, of $L'_{\text{long}}$ and $L'_{\text{lat}}$, from $L_{\text{drift}}$, for localization of the MAV along $Y_\text{W}$ and $X_\text{W}$ axes respectively. Since location along these axes are independent of each other, we treat them separately to solve two 1D localization problems. In the following subsections, without loss of generality, we illustrate our methods on $L'_{\text{long}}$ which can be similarly applied on $L'_{\text{lat}}$.

\begin{figure}[h]
	\def\svgwidth{\columnwidth}{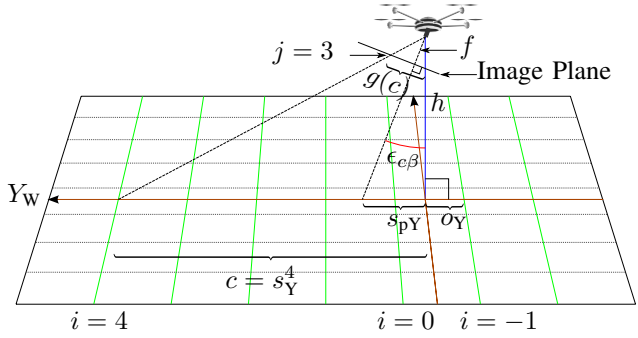}
	\caption{The grid-lines model illustration, showing the MAV state, sub-cell position ($o_\text{Y}$) along $Y_\text{W}$ and projections onto the image. The green lines show the longitudinal lines being considered. The brown lines show the co-ordinate frame on the grid-based floor with origin at the MAV's position on the plane. The blue lines is the perpendicular drawn from the camera's optical center to the grid-based floor. The image plane is exaggerated here for illustration.}
	\label{fig:gridModel}
	\vspace{-0.4cm}
\end{figure}

\subsection{Grid-Lines Model}
We consider the distance ($s_{\text{Y}}$) between each longitudinal line of the grid-based floor and the camera position along $Y_\text{W}$ axis. The line to the immediate positive $Y_\text{W}$ direction, with respect to the camera position is indexed as $i = 0$. Therefore, the unit cell directly bellow the camera consists of lines $i=-1$ and $i=0$. The distance between the camera position and the $i=-1$ line is the sub-cell position of the camera, denoted as $o_{\text{Y}}$ (figure \ref{fig:gridModel}). If $m_\text{Y}$ meters is the size of the unit cell in $Y_\text{W}$ axis, then the domain of $o_{\text{Y}}$ is $[0,m_\text{Y})$.

The distance of $i^{th}$ longitudinal line of the grid-based floor along $Y_\text{W}$ axis with camera position as the reference is given by 
\begin{equation}
	s_{\text{Y}}^i = m_{\text{Y}} \times (i+1) - o_Y
\end{equation}

The normal to the image plane intersects the floor plane at 
\begin{equation}
	s_\text{pY} = h \times tan(\epsilon_{c\beta})
\end{equation}
Here $h$ is the height of the camera from the grid-based floor in meters.

The projection of any line of magnitude $c$ in $Y_\text{W}$ axis with camera position as the origin is given by $g(c)$
\begin{equation}
	g(c) = \frac{c \times cos(\phi_Y) \times f}{cos(\delta) \times h}
\end{equation}
where $\phi_Y = tan^{-1}(c/h)$ and $\delta = \phi_Y - \epsilon_{c\beta}$. \\

Since $L'_\text{long}$ is a set of ordered set of $(\rho', \theta', \l)$, let us denote $\rho^j$ as the $\rho'$ from the ordered set whose $l = j$. The labels $j$ and $i$ belong to the lines in the image and grid-based floor respectively and are related by
\begin{equation}
	i = j + \floor*{\frac{s_\text{pY} - o_{\text{Y}}}{m_\text{Y}}}
\end{equation}

We model the distance ($\rho_\text{mod}^j$) in pixels between $j^\text{th}$ longitudinal line and the image center as
\begin{equation}
	\rho_\text{mod}^j = g(s_{\text{Y}}^i) - g(s_\text{pY})
\end{equation}
Hence, the grid-lines model depends on $\epsilon_{c\beta}$, $i$, $m_\text{Y}$, $o_\text{Y}$, $f$ and $h$, where only $o_\text{Y}$ and $h$ are unknown.


\subsection{Cost Function and Optimization}
We define a cost function to determine the two unknowns $o_\text{Y}$ and $h$, based on the observed $\rho^j$ and modelled $\rho_\text{mod}^j$ as
\begin{equation}
	C = \frac{1}{2}\sum_{j}\left( \lvert\lvert  \rho^j - I_\text{X} - \rho_\text{mod}^j  \rvert\rvert^2 \right)
\end{equation}
Hence, we formulate the 1D localization problem as a Non-Linear Least Squares optimization problem with bound constraints. We solve for $o_\text{Y}$ and $h$ with the bounds $0 < o_Y < s_{\text{Y}}$ and $h > 0$ respectively.

In mLOG, we use Ceres Solver \cite{ceres-solver}, an open source C++ library for modelling and solving large, complicated optimization problems with ``automatic differentiation'' enabled and ``DENSE\_QR'' type of linear solver.  

Let $E_\text{f}^k$ be the final cost of optimization at $k^\text{th}$ frame. The optimization result is accepted if 1) Ceres Solver achieves convergence and 2) if there is no abnormal change in final energy, i.e. $$E_\text{f}^k < E_\text{f}^{k-1} * \epsilon_\text{E}$$
We empirically set $\epsilon_\text{E} = 1.0 \times 10^2$. 

\section{GRID LOCALIZATION}
\label{GridLocalization}
As we have denoted, ($o_\text{X}$, $o_\text{Y}$) is the camera position within the unit cell directly below the camera. As the MAV would move over the grid, the unit cell directly below it, will keep on changing. While $o_\text{X}$, $o_\text{Y}$ would be the position over that dynamic unit cell bounded by $m_\text{X}$ and $m_\text{Y}$ respectively, we integrate the position at every sequential frame to keep track of the unit cells, the MAV has traversed. 

Let $o_\text{Y}^{k}$ be the position within $u_k$ unit cell and $p_\text{Y}^k$ be the position of the MAV with respect to an initial position over the grid-based floor at $k^\text{th}$ frame. Hence we have 
\begin{equation*}
	u_k = \floor*{ \frac{p_\text{Y}^k}{m_{\text{Y}}} }
\end{equation*} 
At $(k+1)^\text{th}$ frame, the new sub-cell position $o_Y^{k+1}$, might be from $u_k-1$, $u_k$ or $u_k+1$ unit cell, considering the maximum MAV speed is limited as discussed in section \ref{ConventionsAndAssumptions} above. Hence three possible position of the MAV at $(k+1)^\text{th}$ frame are
\begin{equation*}
	P_\text{Y} = \{(p_\text{Y}-1),  (p_\text{Y}), (p_\text{Y}+1)\}
\end{equation*}
where $p_\text{Y} = p_\text{Y}^{k}+o_\text{Y}^{k+1}-o_\text{Y}^{k}$. The MAV's new position ($p_\text{Y}^{k+1}$) is given by a winner take all (WTA) scheme, decided by
\begin{equation}
	p_\text{Y}^{k+1} = \underset{p'_\text{Y} \in P_\text{Y}}{arg\,min} (p_\text{Y}^{k} - p'_\text{Y})^2
\end{equation}

Hence we can observe the limitation of $\epsilon_\text{s} \geq 3$, since for $\epsilon_\text{s} < 3$ the position over the grid will become ambiguous in nearby unit cells for some cases and lead to drift errors. 

\begin{table}
\captionof{table}{Thresholds and physical constants}\label{constants}
\centering
\resizebox{0.9\columnwidth}{!}{\begin{tabular}{ | l | l | l | l | }
	\hline
	\textbf{Name} & \textbf{Symbol} & \textbf{Domain} & \textbf{Value} \\
	\hline
	Speed factor & $\epsilon_\text{s}$ & $\geq 3$ & 3\\
	RANSAC width threshold  & $d$ & $>0$ & 0.1 \\
	RANSAC minimum number to assert a fit  & $t$ & $>0$ & 10 \\
	KDE Bandwidth & $b$ & $>0$ & 20 \\
	Clustering threshold & $\epsilon_\text{c}$ & $>0$ & $max \hat{f_b}(\rho) \times 0.25$ \\
	Energy threshold & $\epsilon_\text{E}$ & $\mathbb{R}$ & $1.0 \times 10^2$ \\
	\hline
\end{tabular}}
\vspace{-0.5cm}
\end{table}

\section{EXPERIMENTAL RESULTS}
\label{exprerimentalResults}
We performed a set of experiments to evaluate 1) the localization accuracy of the proposed grid-based method, and, 2) correlation between the errors in state variables.

\begin{figure}[h]
	\vspace{-0.2cm}
	\subfloat{\includegraphics[width=0.5\columnwidth]{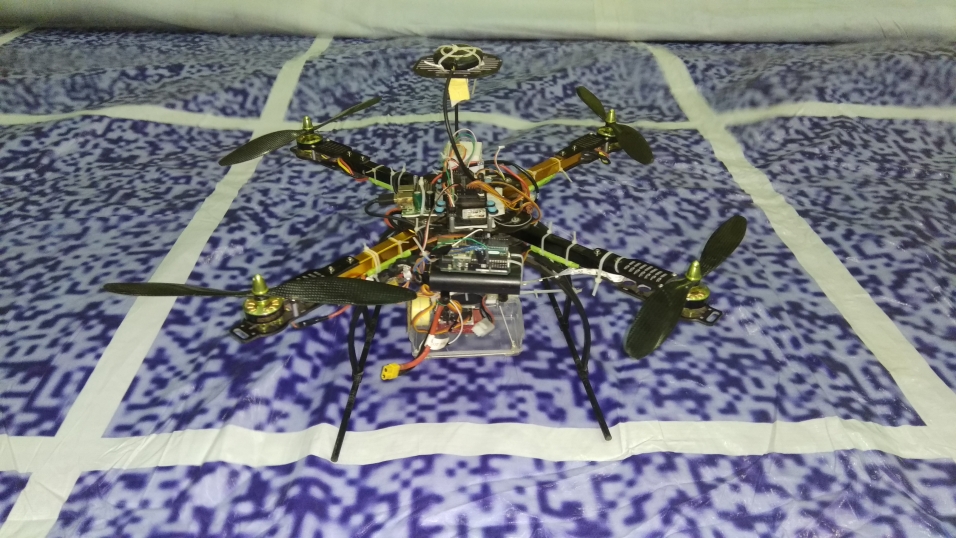}}
	\subfloat{\includegraphics[width=0.5\columnwidth]{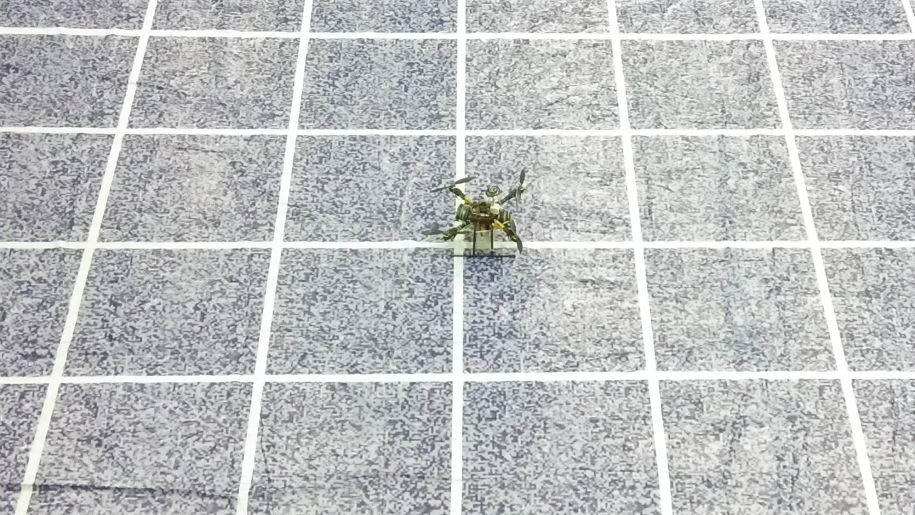}}
	\caption{\textbf{Left}: X650F quadcopter. \textbf{Right}: Testing Arena}
	\label{fig:picQuadArena}
	\vspace{-0.2cm}
\end{figure}

The experiments were performed on a 650 size quadcopter, with a downward facing Logitech B910 HD webcam (figure \ref{fig:picQuadArena}). The localization was performed in real-time at 30 Hz on a Raspberry Pi3 onboard computer with parameters as shown in table \ref{constants}. Except the speed factor ($\epsilon_\text{s}$), that is set by design of the proposed method, all the other parameters are set empirically and related generally on the size of the unit grid cell. The flight control board is based on ArduPilot Mega and is customized to accept position feedback from motion capture system. A $10\text{m} \times 10\text{m}$ grid of square unit cell dimension $s_\text{X} = s_\text{Y} = 1\text{m}$ was printed on flex to form the grid-based floor. The grid-based floor resembles the arena of IARC mission 7. Motion capture system is used for the ground truth localization of the quadcopter. The image resolution was $640 \times 480$ pixels.

\begin{figure}[h!]
	\vspace{-0.2cm}
	\subfloat{\includegraphics[width=\columnwidth]{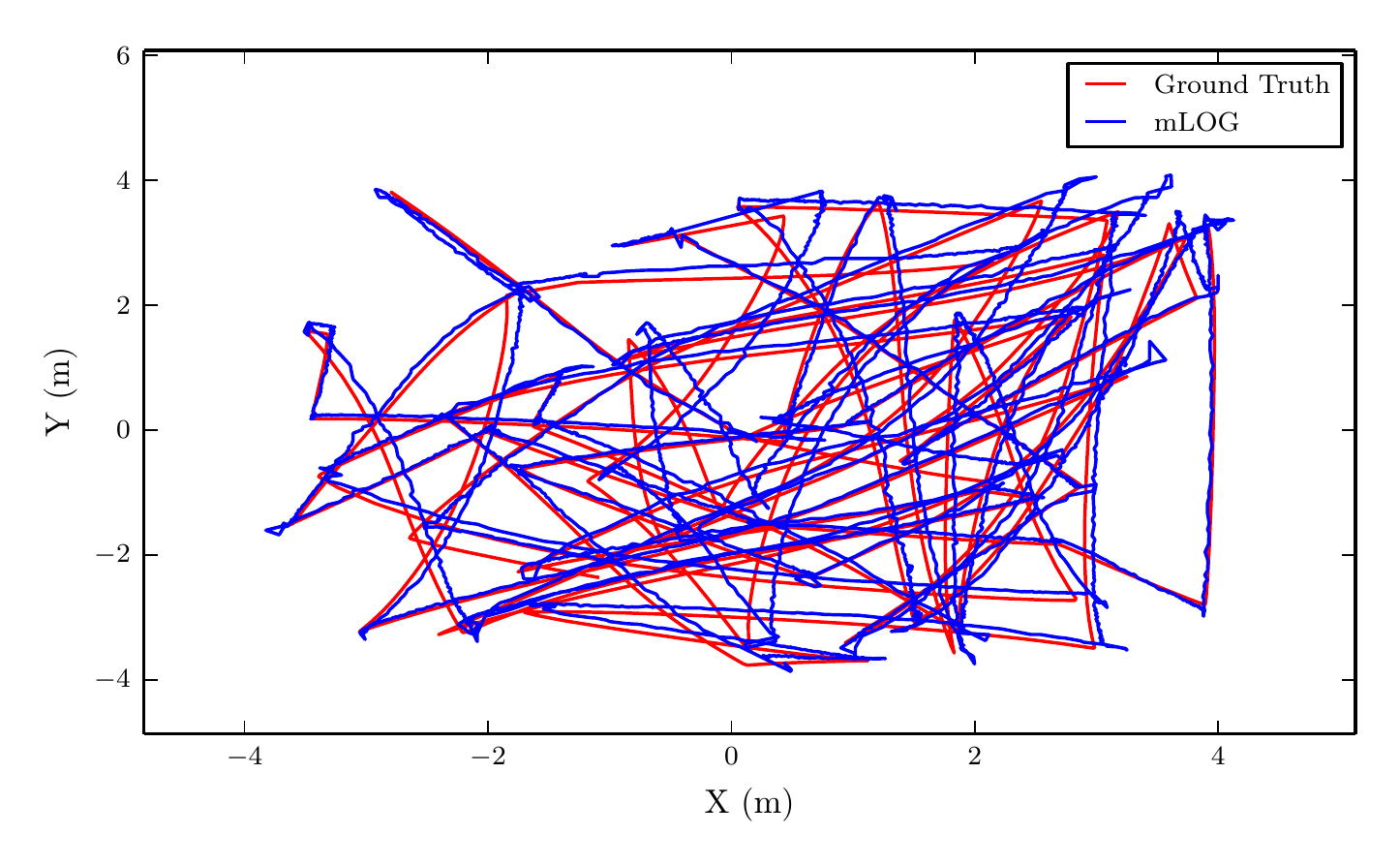}}
	\vspace{-0.2cm}
	\subfloat{\includegraphics[width=0.5\columnwidth]{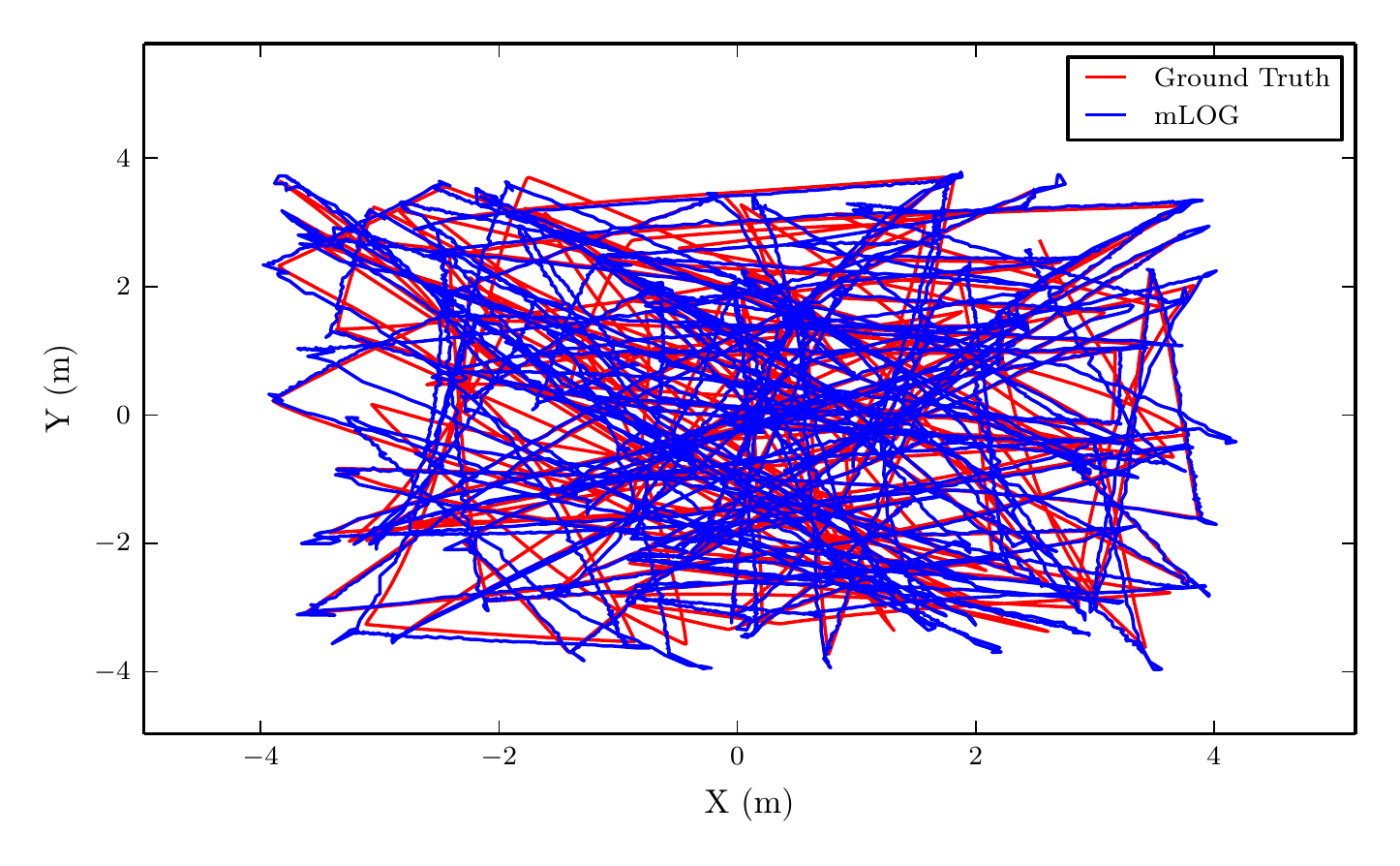}}
	\subfloat{\includegraphics[width=0.5\columnwidth]{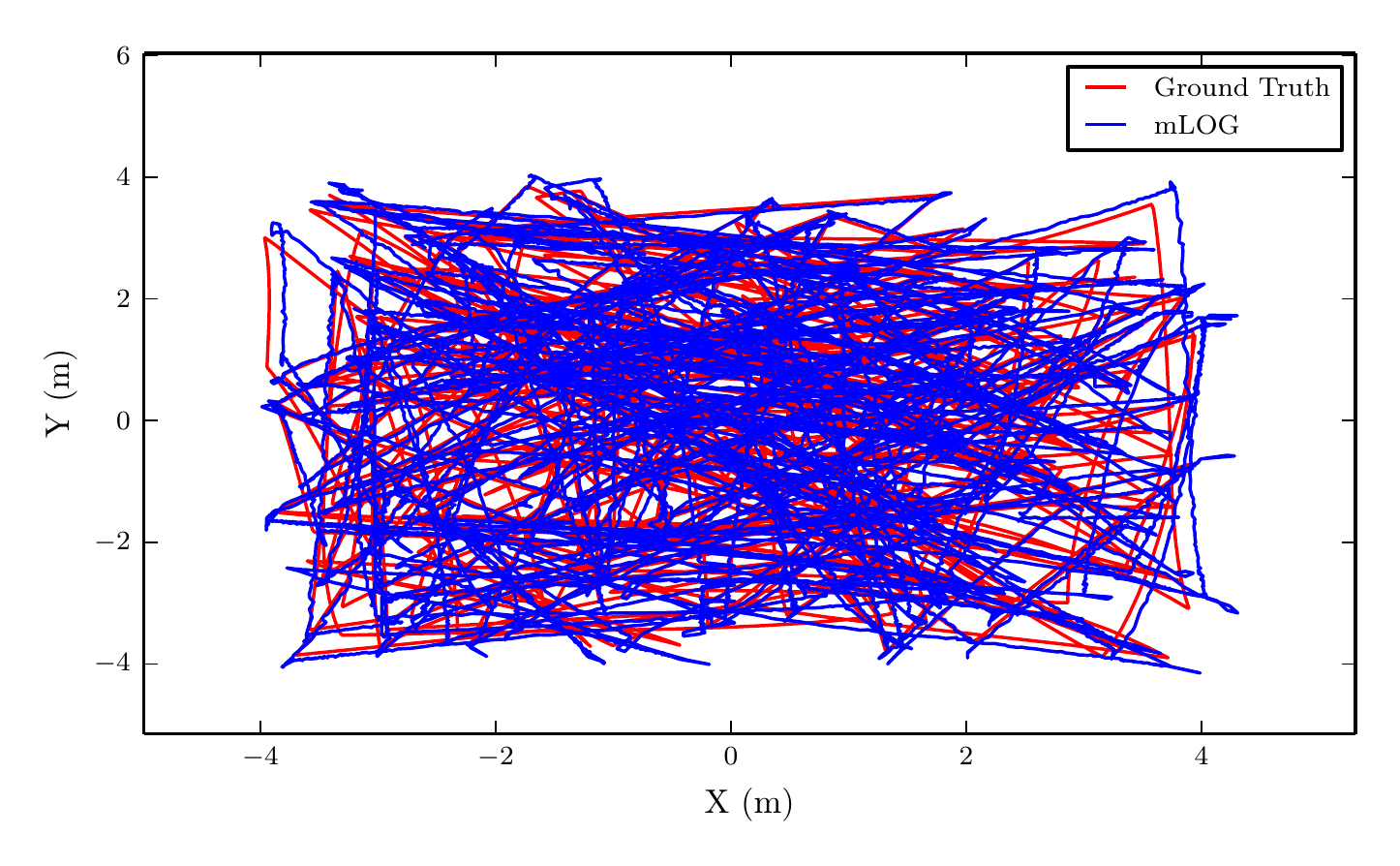}}
	\vspace{-0.2cm}
	\subfloat{\includegraphics[width=\columnwidth]{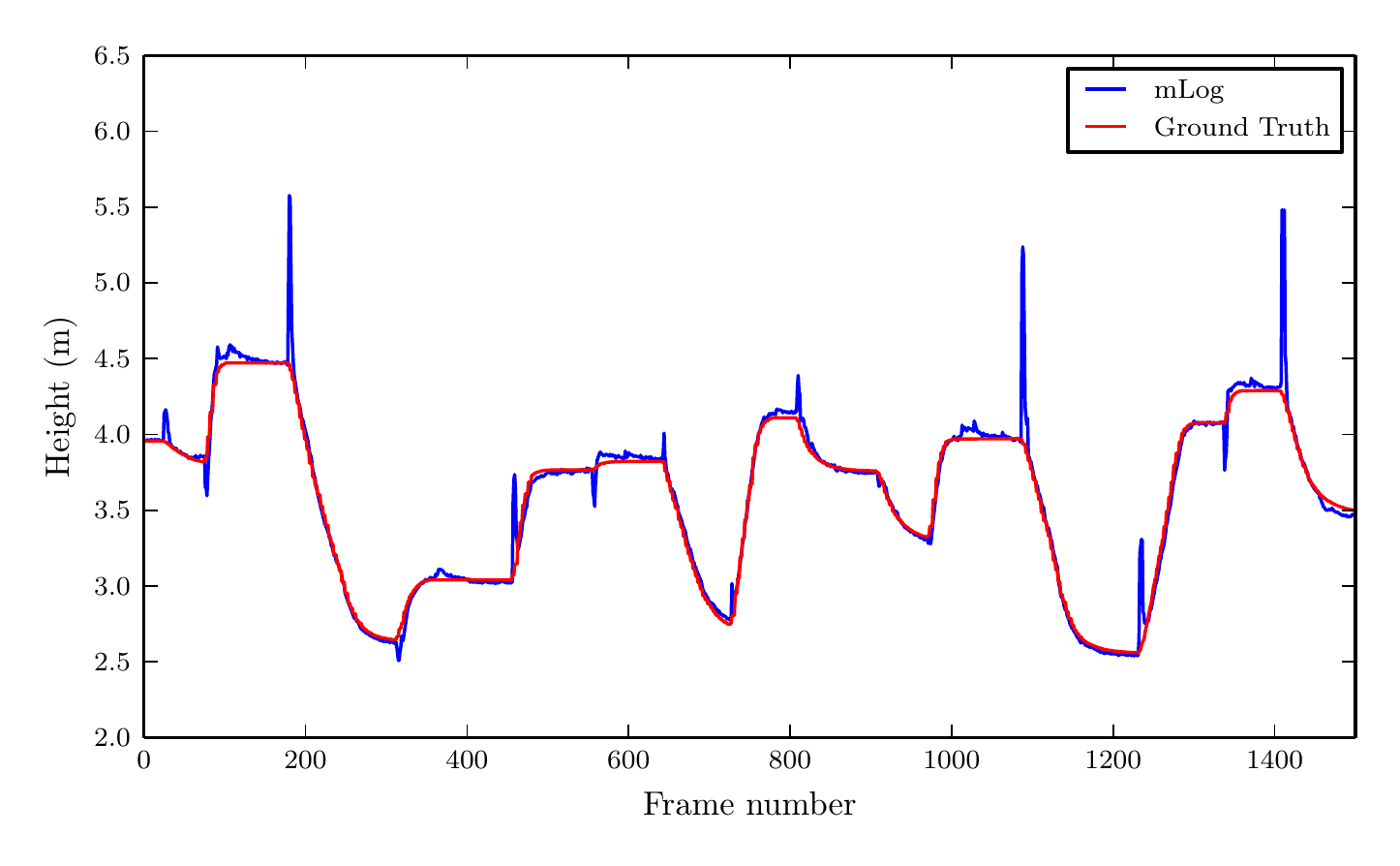}}
	\vspace{-0.2cm}
	\caption{Ground truth (Red) vs mLOG (Blue) localization with path lengths 264.60m (\textbf{Top}), 639.34m (\textbf{Mid left}) and 1020.73m (\textbf{Mid right}), and sample height estimate (\textbf{Bottom}). Units are in meters.}
	\label{fig:picXYPlot}
\end{figure}

Three trial flights (figure \ref{fig:picXYPlot}) were made of varying path lengths within speed limits (section \ref{ConventionsAndAssumptions}). Note that the proposed method would fail if operated beyond the speed limit. The set-points for the paths were generated randomly within the arena bounds, and the quadcopter was flown autonomously. The three paths were of length 264.60m, 639.34m and 1020.73m respectively. We chose to conduct multiple flights in order to demonstrate reliability and performance of the proposed localization method with increasing path length.

\subsection{Localization Accuracy}
The mean error, root-mean-square error (RMSE) and error standard deviation (SD) in the state variables for the three paths are reported in Table \ref{table:tableError250}, \ref{table:tableError500} and \ref{table:tableError1000} respectively. Error in X, Y and Z state variables are reported in meters, while error in pitch and roll are reported in degrees. The average standard deviation in errors are 0.14m, 0.13m, 0.17m, 1.06$^{\circ}$ and 1.06$^{\circ}$ for X, Y, Z, pitch and roll respectively.

\begin{table}[h!]
\centering
\begin{tabular}{ |c|ccccc| } 
 \hline
  & X & Y & Z & Pitch & Roll \\
  \hline
  Mean & 0.12 & -0.09 & 0.021 & 0.03 & 0.05 \\
  RMSE & 0.17 & 0.14 & 0.11 & 1.03 & 1.06 \\
  SD & 0.11 & 0.10 & 0.10 & 1.03 & 1.06 \\
 \hline
\end{tabular}
\caption{For path length 264.60m}
\label{table:tableError250}
\vspace{-0.2cm}
\end{table}

\begin{table}[h!]
\centering
\begin{tabular}{ |c|ccccc| } 
 \hline
  & X & Y & Z & Pitch & Roll \\
  \hline
  Mean & -0.02 & -0.03 & 0.02 & 0.06 & 0.03 \\
  RMSE & 0.13 & 0.14 & 0.32 & 1.07 & 1.03 \\
  SD & 0.14 & 0.14 & 0.31 & 1.07 & 1.03 \\
 \hline
\end{tabular}
\caption{For path length 639.34m}
\label{table:tableError500}
\vspace{-0.2cm}
\end{table}

\begin{table}[h!]
\centering
\begin{tabular}{ |c|ccccc| } 
 \hline
  & X & Y & Z & Pitch & Roll \\
  \hline
  Mean & 0.04 & -0.05 & 0.02 & 0.05 & 0.04 \\
  RMSE & 0.17 & 0.14 & 0.10 & 1.08 & 1.10 \\
  SD & 0.16 & 0.14 & 0.10 & 1.08 & 1.10 \\
 \hline
\end{tabular}
\caption{For path length 1020.73m}
\label{table:tableError1000}
\vspace{-0.4cm}
\end{table}

\subsection{Correlation in errors between state variables}
In the proposed method, since estimation of some state variables depend on the estimation of others, hence we were interested in finding the correlation in errors between the state variables. We expected positive correlation between errors in orientation and position estimation as the estimation of later is dependent on the former (section \ref{orientationAndCompensation}). The correlations for the three flights are reported in the table \ref{table:tableCorr250}, \ref{table:tableCorr500} and \ref{table:tableCorr1000}.

\begin{table}[h!]
\centering
\begin{tabular}{ |c|ccccc| } 
 \hline
  & X & Y & Z & Pitch & Roll \\
  \hline
  X & 1.0 & -0.75 & 0.05 & -0.05 & 0.03 \\
  Y & -0.75 & 1.0 & -0.02 & 0.05 & -0.01 \\
  Z & 0.05 & -0.02 & 1.0 & 0.14 & 0.25 \\
  Pitch & -0.05 & 0.05 & 0.14 & 1.0 & -0.17 \\
  Roll & 0.03 & -0.01 & 0.25 & -0.17 & 1.0 \\
 \hline
\end{tabular}
\caption{For path length 264.60m}
\label{table:tableCorr250}
\vspace{-0.2cm}
\end{table}

\begin{table}[h!]
\centering
\begin{tabular}{ |c|ccccc| } 
 \hline
  & X & Y & Z & Pitch & Roll \\
  \hline
  X & 1.0 & -0.70 & 0.00 & -0.02 & 0.06 \\
  Y & -0.70 & 1.0 & 0.02 & 0.03 & -0.02 \\
  Z & 0.00 & 0.02 & 1.0 & 0.08 & 0.07 \\
  Pitch & -0.02 & 0.03 & 0.08 & 1.0 & 0.14 \\
  Roll & 0.06 & -0.02 & 0.07 & 0.14 & 1.0 \\
 \hline
\end{tabular}
\caption{For path length 639.34m}
\label{table:tableCorr500}
\vspace{-0.2cm}
\end{table}

\begin{table}[h!]
\centering
\begin{tabular}{ |c|ccccc| } 
 \hline
  & X & Y & Z & Pitch & Roll \\
  \hline
  X & 1.0 & -0.79 & 0.02 & -0.03 & 0.04 \\
  Y & -0.79 & 1.0 & 0.03 & 0.05 & -0.01 \\
  Z & 0.02 & 0.03 & 1.0 & 0.19 & 0.22 \\
  Pitch & -0.03 & 0.05 & 0.19 & 1.0 & 0.04 \\
  Roll & 0.04 & -0.01 & 0.22 & 0.04 & 1.0 \\
 \hline
\end{tabular}
\caption{For path length 1020.73m}
\label{table:tableCorr1000}
\vspace{-0.4cm}
\end{table}

An interesting, negative correlation between X and Y has been observed, regardless of them being estimated independently. The reason for it might be specific to the quadcopter's motion, that led to the increase in negative Y error when X error increased in the positive direction and vice versa. Further, error in orientation is observed to have a small positive correlation with the error in Z only and not X or Y. Extreme roll or pitch ($\sim$35$^{\circ}$), however, causes an instantaneous error in the height estimate as shown in figure \ref{fig:picXYPlot} (Bottom), in contrast, it does not seem to affect the X and Y estimates.

\section{CONCLUSION AND FURTHER WORK}
In this paper, we introduced a novel algorithm for monocular visual, 5DoF localization of MAVs over a grid-based floor. We demonstrated experimentally that mLOG provides an accurate localization with bounded drift errors. We also showed that the localization can be performed in real-time and onboard the MAV in around 30 Hz. The proposed method advances in a deterministic approach on the detected set of lines. Hence a promising direction for future work would be to use a probabilistic approach to fill lines missing in the detection and modify the gridlines model to enable different camera tilt angle for the perception of more number of grid lines at lower MAV height.

\bibliographystyle{IEEEtran}
\bibliography{ref}

\end{document}